\documentstyle[breakcites,cgloss4e,gb4e,nllt_labels]{article}
\makeatletter
\newcommand{\citedelimiters}[2]{    % changes opening and closing brackets
\def\@cite##1##2{#1{##1\if@tempswa{:}{##2}\fi}#2} % gives colon before page refs
\def\@biblabel##1{#1##1#2\hfill}
}
\makeatother
\citedelimiters{}{}	% gives no delimiters
\makeatletter

\newcommand{\nobiblabels}{\def\@biblabel##1{}
 \@ifundefined{chapter}{\def\thebibliography##1{\section*{References\markboth
  {REFERENCES}{REFERENCES}}\list
  {[\arabic{enumi}]}{\setlength{\labelwidth}{1.5em}
    \setlength{\labelsep}{0em}
    \leftmargin\labelwidth
    \advance\leftmargin\labelsep
    \setlength{\itemindent}{-1.5em}
    \setlength{\itemsep}{0em}
    \setlength{\parsep}{0em}
    \usecounter{enumi}}
    \def\newblock{\hskip .11em plus .33em minus -.07em}
    \sloppy\clubpenalty4000\widowpenalty4000
    \sfcode`\.=1000\relax}}%
{\def\thebibliography##1{\chapter*{Bibliography\markboth
  {BIBLIOGRAPHY}{BIBLIOGRAPHY}}\list
  {[\arabic{enumi}]}{\setlength{\labelwidth}{1.5em}
    \setlength{\labelsep}{0em}
    \leftmargin\labelwidth
    \advance\leftmargin\labelsep
    \setlength{\itemindent}{-1.5em}
    \usecounter{enumi}}
    \def\newblock{\hskip .11em plus .33em minus -.07em}
    \sloppy\clubpenalty4000\widowpenalty4000
    \sfcode`\.=1000\relax}}}

\makeatother

\newcommand{\ems}[1]{\em #1\/}

\newcommand{\bi}{\begin{itemize}}
\newcommand{\ei}{\end{itemize}}

\newcommand{\bt}{\begin{tabbing}}
\newcommand{\et}{\end{tabbing}}

\newcommand{\bex}{\begin{exe}[(999)]}
\newcommand{\eex}{\end{exe}}

\newcommand{\bxl}{\begin{xlist}}
\newcommand{\exl}{\end{xlist}}

\newcommand{\nn}{\noindent}

\nobiblabels

\newcounter{treecount}
\newcounter{branchcount}
\newsavebox{\parentbox}
\newsavebox{\treebox}
\newsavebox{\treeboxone}
\newsavebox{\treeboxtwo}
\newsavebox{\treeboxthree}
\newsavebox{\treeboxfour}
\newsavebox{\treeboxfive}
\newsavebox{\treeboxsix}
\newsavebox{\treeboxseven}
\newsavebox{\treeboxeight}
\newsavebox{\treeboxnine}
\newsavebox{\treeboxten}
\newsavebox{\treeboxeleven}
\newsavebox{\treeboxtwelve}
\newsavebox{\treeboxthirteen}
\newsavebox{\treeboxfourteen}
\newsavebox{\treeboxfifteen}
\newsavebox{\treeboxsixteen}
\newsavebox{\treeboxseventeen}
\newsavebox{\treeboxeighteen}
\newsavebox{\treeboxnineteen}
\newsavebox{\treeboxtwenty}
\newlength{\treeoffsetone}
\newlength{\treeoffsettwo}
\newlength{\treeoffsetthree}
\newlength{\treeoffsetfour}
\newlength{\treeoffsetfive}
\newlength{\treeoffsetsix}
\newlength{\treeoffsetseven}
\newlength{\treeoffseteight}
\newlength{\treeoffsetnine}
\newlength{\treeoffsetten}
\newlength{\treeoffseteleven}
\newlength{\treeoffsettwelve}
\newlength{\treeoffsetthirteen}
\newlength{\treeoffsetfourteen}
\newlength{\treeoffsetfifteen}
\newlength{\treeoffsetsixteen}
\newlength{\treeoffsetseventeen}
\newlength{\treeoffseteighteen}
\newlength{\treeoffsetnineteen}
\newlength{\treeoffsettwenty}

\newlength{\treeshiftone}
\newlength{\treeshifttwo}
\newlength{\treeshiftthree}
\newlength{\treeshiftfour}
\newlength{\treeshiftfive}
\newlength{\treeshiftsix}
\newlength{\treeshiftseven}
\newlength{\treeshifteight}
\newlength{\treeshiftnine}
\newlength{\treeshiftten}
\newlength{\treeshifteleven}
\newlength{\treeshifttwelve}
\newlength{\treeshiftthirteen}
\newlength{\treeshiftfourteen}
\newlength{\treeshiftfifteen}
\newlength{\treeshiftsixteen}
\newlength{\treeshiftseventeen}
\newlength{\treeshifteighteen}
\newlength{\treeshiftnineteen}
\newlength{\treeshifttwenty}
\newlength{\treewidthone}
\newlength{\treewidthtwo}
\newlength{\treewidththree}
\newlength{\treewidthfour}
\newlength{\treewidthfive}
\newlength{\treewidthsix}
\newlength{\treewidthseven}
\newlength{\treewidtheight}
\newlength{\treewidthnine}
\newlength{\treewidthten}
\newlength{\treewidtheleven}
\newlength{\treewidthtwelve}
\newlength{\treewidththirteen}
\newlength{\treewidthfourteen}
\newlength{\treewidthfifteen}
\newlength{\treewidthsixteen}
\newlength{\treewidthseventeen}
\newlength{\treewidtheighteen}
\newlength{\treewidthnineteen}
\newlength{\treewidthtwenty}
\newlength{\daughteroffsetone}
\newlength{\daughteroffsettwo}
\newlength{\daughteroffsetthree}
\newlength{\daughteroffsetfour}
\newlength{\branchwidthone}
\newlength{\branchwidthtwo}
\newlength{\branchwidththree}
\newlength{\branchwidthfour}
\newlength{\parentoffset}
\newlength{\treeoffset}
\newlength{\daughteroffset}
\newlength{\branchwidth}
\newlength{\parentwidth}
\newlength{\treewidth}
\newcommand{\ontop}[1]{\begin{tabular}{c}#1\end{tabular}}
\newcommand{\poptree}{%
\ifnum\value{treecount}=0\typeout{QobiTeX warning---Tree stack underflow}\fi%
\addtocounter{treecount}{-1}%
\setlength{\treeoffsettwo}{\treeoffsetthree}%
\setlength{\treeoffsetthree}{\treeoffsetfour}%
\setlength{\treeoffsetfour}{\treeoffsetfive}%
\setlength{\treeoffsetfive}{\treeoffsetsix}%
\setlength{\treeoffsetsix}{\treeoffsetseven}%
\setlength{\treeoffsetseven}{\treeoffseteight}%
\setlength{\treeoffseteight}{\treeoffsetnine}%
\setlength{\treeoffsetnine}{\treeoffsetten}%
\setlength{\treeoffsetten}{\treeoffseteleven}%
\setlength{\treeoffseteleven}{\treeoffsettwelve}%
\setlength{\treeoffsettwelve}{\treeoffsetthirteen}%
\setlength{\treeoffsetthirteen}{\treeoffsetfourteen}%
\setlength{\treeoffsetfourteen}{\treeoffsetfifteen}%
\setlength{\treeoffsetfifteen}{\treeoffsetsixteen}%
\setlength{\treeoffsetsixteen}{\treeoffsetseventeen}%
\setlength{\treeoffsetseventeen}{\treeoffseteighteen}%
\setlength{\treeoffseteighteen}{\treeoffsetnineteen}%
\setlength{\treeoffsetnineteen}{\treeoffsettwenty}%
\setlength{\treeshifttwo}{\treeshiftthree}%
\setlength{\treeshiftthree}{\treeshiftfour}%
\setlength{\treeshiftfour}{\treeshiftfive}%
\setlength{\treeshiftfive}{\treeshiftsix}%
\setlength{\treeshiftsix}{\treeshiftseven}%
\setlength{\treeshiftseven}{\treeshifteight}%
\setlength{\treeshifteight}{\treeshiftnine}%
\setlength{\treeshiftnine}{\treeshiftten}%
\setlength{\treeshiftten}{\treeshifteleven}%
\setlength{\treeshifteleven}{\treeshifttwelve}%
\setlength{\treeshifttwelve}{\treeshiftthirteen}%
\setlength{\treeshiftthirteen}{\treeshiftfourteen}%
\setlength{\treeshiftfourteen}{\treeshiftfifteen}%
\setlength{\treeshiftfifteen}{\treeshiftsixteen}%
\setlength{\treeshiftsixteen}{\treeshiftseventeen}%
\setlength{\treeshiftseventeen}{\treeshifteighteen}%
\setlength{\treeshifteighteen}{\treeshiftnineteen}%
\setlength{\treeshiftnineteen}{\treeshifttwenty}%
\setlength{\treewidthtwo}{\treewidththree}%
\setlength{\treewidththree}{\treewidthfour}%
\setlength{\treewidthfour}{\treewidthfive}%
\setlength{\treewidthfive}{\treewidthsix}%
\setlength{\treewidthsix}{\treewidthseven}%
\setlength{\treewidthseven}{\treewidtheight}%
\setlength{\treewidtheight}{\treewidthnine}%
\setlength{\treewidthnine}{\treewidthten}%
\setlength{\treewidthten}{\treewidtheleven}%
\setlength{\treewidtheleven}{\treewidthtwelve}%
\setlength{\treewidthtwelve}{\treewidththirteen}%
\setlength{\treewidththirteen}{\treewidthfourteen}%
\setlength{\treewidthfourteen}{\treewidthfifteen}%
\setlength{\treewidthfifteen}{\treewidthsixteen}%
\setlength{\treewidthsixteen}{\treewidthseventeen}%
\setlength{\treewidthseventeen}{\treewidtheighteen}%
\setlength{\treewidtheighteen}{\treewidthnineteen}%
\setlength{\treewidthnineteen}{\treewidthtwenty}%
\sbox{\treeboxtwo}{\usebox{\treeboxthree}}%
\sbox{\treeboxthree}{\usebox{\treeboxfour}}%
\sbox{\treeboxfour}{\usebox{\treeboxfive}}%
\sbox{\treeboxfive}{\usebox{\treeboxsix}}%
\sbox{\treeboxsix}{\usebox{\treeboxseven}}%
\sbox{\treeboxseven}{\usebox{\treeboxeight}}%
\sbox{\treeboxeight}{\usebox{\treeboxnine}}%
\sbox{\treeboxnine}{\usebox{\treeboxten}}%
\sbox{\treeboxten}{\usebox{\treeboxeleven}}%
\sbox{\treeboxeleven}{\usebox{\treeboxtwelve}}%
\sbox{\treeboxtwelve}{\usebox{\treeboxthirteen}}%
\sbox{\treeboxthirteen}{\usebox{\treeboxfourteen}}%
\sbox{\treeboxfourteen}{\usebox{\treeboxfifteen}}%
\sbox{\treeboxfifteen}{\usebox{\treeboxsixteen}}%
\sbox{\treeboxsixteen}{\usebox{\treeboxseventeen}}%
\sbox{\treeboxseventeen}{\usebox{\treeboxeighteen}}%
\sbox{\treeboxeighteen}{\usebox{\treeboxnineteen}}%
\sbox{\treeboxnineteen}{\usebox{\treeboxtwenty}}}
\newcommand{\leaf}[1]{%
\ifnum\value{treecount}=20\typeout{QobiTeX warning---Tree stack overflow}\fi%
\addtocounter{treecount}{1}%
\sbox{\treeboxtwenty}{\usebox{\treeboxnineteen}}%
\sbox{\treeboxnineteen}{\usebox{\treeboxeighteen}}%
\sbox{\treeboxeighteen}{\usebox{\treeboxseventeen}}%
\sbox{\treeboxseventeen}{\usebox{\treeboxsixteen}}%
\sbox{\treeboxsixteen}{\usebox{\treeboxfifteen}}%
\sbox{\treeboxfifteen}{\usebox{\treeboxfourteen}}%
\sbox{\treeboxfourteen}{\usebox{\treeboxthirteen}}%
\sbox{\treeboxthirteen}{\usebox{\treeboxtwelve}}%
\sbox{\treeboxtwelve}{\usebox{\treeboxeleven}}%
\sbox{\treeboxeleven}{\usebox{\treeboxten}}%
\sbox{\treeboxten}{\usebox{\treeboxnine}}%
\sbox{\treeboxnine}{\usebox{\treeboxeight}}%
\sbox{\treeboxeight}{\usebox{\treeboxseven}}%
\sbox{\treeboxseven}{\usebox{\treeboxsix}}%
\sbox{\treeboxsix}{\usebox{\treeboxfive}}%
\sbox{\treeboxfive}{\usebox{\treeboxfour}}%
\sbox{\treeboxfour}{\usebox{\treeboxthree}}%
\sbox{\treeboxthree}{\usebox{\treeboxtwo}}%
\sbox{\treeboxtwo}{\usebox{\treeboxone}}%
\sbox{\treeboxone}{\ontop{#1}}%
\sbox{\treeboxone}{\raisebox{-\ht\treeboxone}{\usebox{\treeboxone}}}%
\setlength{\treeoffsettwenty}{\treeoffsetnineteen}%
\setlength{\treeoffsetnineteen}{\treeoffseteighteen}%
\setlength{\treeoffseteighteen}{\treeoffsetseventeen}%
\setlength{\treeoffsetseventeen}{\treeoffsetsixteen}%
\setlength{\treeoffsetsixteen}{\treeoffsetfifteen}%
\setlength{\treeoffsetfifteen}{\treeoffsetfourteen}%
\setlength{\treeoffsetfourteen}{\treeoffsetthirteen}%
\setlength{\treeoffsetthirteen}{\treeoffsettwelve}%
\setlength{\treeoffsettwelve}{\treeoffseteleven}%
\setlength{\treeoffseteleven}{\treeoffsetten}%
\setlength{\treeoffsetten}{\treeoffsetnine}%
\setlength{\treeoffsetnine}{\treeoffseteight}%
\setlength{\treeoffseteight}{\treeoffsetseven}%
\setlength{\treeoffsetseven}{\treeoffsetsix}%
\setlength{\treeoffsetsix}{\treeoffsetfive}%
\setlength{\treeoffsetfive}{\treeoffsetfour}%
\setlength{\treeoffsetfour}{\treeoffsetthree}%
\setlength{\treeoffsetthree}{\treeoffsettwo}%
\setlength{\treeoffsettwo}{\treeoffsetone}%
\setlength{\treeoffsetone}{0.5\wd\treeboxone}%
\setlength{\treeshifttwenty}{\treeshiftnineteen}%
\setlength{\treeshiftnineteen}{\treeshifteighteen}%
\setlength{\treeshifteighteen}{\treeshiftseventeen}%
\setlength{\treeshiftseventeen}{\treeshiftsixteen}%
\setlength{\treeshiftsixteen}{\treeshiftfifteen}%
\setlength{\treeshiftfifteen}{\treeshiftfourteen}%
\setlength{\treeshiftfourteen}{\treeshiftthirteen}%
\setlength{\treeshiftthirteen}{\treeshifttwelve}%
\setlength{\treeshifttwelve}{\treeshifteleven}%
\setlength{\treeshifteleven}{\treeshiftten}%
\setlength{\treeshiftten}{\treeshiftnine}%
\setlength{\treeshiftnine}{\treeshifteight}%
\setlength{\treeshifteight}{\treeshiftseven}%
\setlength{\treeshiftseven}{\treeshiftsix}%
\setlength{\treeshiftsix}{\treeshiftfive}%
\setlength{\treeshiftfive}{\treeshiftfour}%
\setlength{\treeshiftfour}{\treeshiftthree}%
\setlength{\treeshiftthree}{\treeshifttwo}%
\setlength{\treeshifttwo}{\treeshiftone}%
\setlength{\treeshiftone}{0pt}%
\setlength{\treewidthtwenty}{\treewidthnineteen}%
\setlength{\treewidthnineteen}{\treewidtheighteen}%
\setlength{\treewidtheighteen}{\treewidthseventeen}%
\setlength{\treewidthseventeen}{\treewidthsixteen}%
\setlength{\treewidthsixteen}{\treewidthfifteen}%
\setlength{\treewidthfifteen}{\treewidthfourteen}%
\setlength{\treewidthfourteen}{\treewidththirteen}%
\setlength{\treewidththirteen}{\treewidthtwelve}%
\setlength{\treewidthtwelve}{\treewidtheleven}%
\setlength{\treewidtheleven}{\treewidthten}%
\setlength{\treewidthten}{\treewidthnine}%
\setlength{\treewidthnine}{\treewidtheight}%
\setlength{\treewidtheight}{\treewidthseven}%
\setlength{\treewidthseven}{\treewidthsix}%
\setlength{\treewidthsix}{\treewidthfive}%
\setlength{\treewidthfive}{\treewidthfour}%
\setlength{\treewidthfour}{\treewidththree}%
\setlength{\treewidththree}{\treewidthtwo}%
\setlength{\treewidthtwo}{\treewidthone}%
\setlength{\treewidthone}{\wd\treeboxone}}
\newcommand{\branch}[2]{%
\setcounter{branchcount}{#1}%
\ifnum\value{branchcount}=1\sbox{\parentbox}{\ontop{#2}}%
\setlength{\parentoffset}{\treeoffsetone}%
\addtolength{\parentoffset}{-0.5\wd\parentbox}%
\setlength{\daughteroffset}{0in}%
\ifdim\parentoffset<0in%
\setlength{\daughteroffset}{-\parentoffset}%
\setlength{\parentoffset}{0in}\fi%
\setlength{\parentwidth}{\parentoffset}%
\addtolength{\parentwidth}{\wd\parentbox}%
\setlength{\treeoffset}{\daughteroffset}%
\addtolength{\treeoffset}{\treeoffsetone}%
\setlength{\treewidth}{\wd\treeboxone}%
\addtolength{\treewidth}{\daughteroffset}%
\ifdim\treewidth<\parentwidth\setlength{\treewidth}{\parentwidth}\fi%
\sbox{\treebox}{\begin{minipage}{\treewidth}%
\begin{flushleft}%
\hspace*{\parentoffset}\usebox{\parentbox}\\
{\setlength{\unitlength}{2ex}%
\hspace*{\treeoffset}\begin{picture}(0,1)%
\put(0,0){\line(0,1){1}}%
\end{picture}}\\
\vspace{-\baselineskip}
\hspace*{\daughteroffset}%
\raisebox{-\ht\treeboxone}{\usebox{\treeboxone}}%
\end{flushleft}%
\end{minipage}}%
\setlength{\treeoffsetone}{\parentoffset}%
\addtolength{\treeoffsetone}{0.5\wd\parentbox}%
\setlength{\treeshiftone}{0pt}%
\setlength{\treewidthone}{\treewidth}%
\sbox{\treeboxone}{\usebox{\treebox}}%
\else\ifnum\value{branchcount}=2\sbox{\parentbox}{\ontop{#2}}%
\setlength{\branchwidthone}{\treewidthtwo}%
\addtolength{\branchwidthone}{\treeoffsetone}%
\addtolength{\branchwidthone}{-\treeshiftone}%
\addtolength{\branchwidthone}{-\treeoffsettwo}%
\setlength{\branchwidth}{\branchwidthone}%
\setlength{\daughteroffsetone}{\branchwidth}%
\addtolength{\daughteroffsetone}{-\branchwidthone}%
\addtolength{\daughteroffsetone}{-\treeshiftone}%
\setlength{\parentoffset}{-0.5\wd\parentbox}%
\addtolength{\parentoffset}{\treeoffsettwo}%
\addtolength{\parentoffset}{0.5\branchwidth}%
\setlength{\daughteroffset}{0in}%
\ifdim\parentoffset<0in%
\setlength{\daughteroffset}{-\parentoffset}%
\setlength{\parentoffset}{0in}\fi%
\setlength{\parentwidth}{\parentoffset}%
\addtolength{\parentwidth}{\wd\parentbox}%
\setlength{\treeoffset}{\daughteroffset}%
\addtolength{\treeoffset}{\treeoffsettwo}%
\setlength{\treewidth}{\wd\treeboxone}%
\addtolength{\treewidth}{\daughteroffsetone}%
\addtolength{\treewidth}{\treewidthtwo}%
\addtolength{\treewidth}{\daughteroffset}%
\ifdim\treewidth<\parentwidth\setlength{\treewidth}{\parentwidth}\fi%
\sbox{\treebox}{\begin{minipage}{\treewidth}%
\begin{flushleft}%
\hspace*{\parentoffset}\usebox{\parentbox}\\
{\setlength{\unitlength}{0.5\branchwidth}%
\hspace*{\treeoffset}\begin{picture}(2,0.5)%
\put(0,0){\line(2,1){1}}%
\put(2,0){\line(-2,1){1}}%
\end{picture}}\\
\vspace{-\baselineskip}
\hspace*{\daughteroffset}%
\makebox[\treewidthtwo][l]%
{\raisebox{-\ht\treeboxtwo}{\usebox{\treeboxtwo}}}%
\hspace*{\daughteroffsetone}%
\raisebox{-\ht\treeboxone}{\usebox{\treeboxone}}%
\end{flushleft}%
\end{minipage}}%
\setlength{\treeoffsetone}{\parentoffset}%
\addtolength{\treeoffsetone}{0.5\wd\parentbox}%
\setlength{\treeshiftone}{0pt}%
\setlength{\treewidthone}{\treewidth}%
\sbox{\treeboxone}{\usebox{\treebox}}\poptree%
\else\ifnum\value{branchcount}=3\sbox{\parentbox}{\ontop{#2}}%
\setlength{\branchwidthone}{\treewidthtwo}%
\addtolength{\branchwidthone}{\treeoffsetone}%
\addtolength{\branchwidthone}{-\treeshiftone}%
\addtolength{\branchwidthone}{-\treeoffsettwo}%
\setlength{\branchwidthtwo}{\treewidththree}%
\addtolength{\branchwidthtwo}{\treeoffsettwo}%
\addtolength{\branchwidthtwo}{-\treeshifttwo}%
\addtolength{\branchwidthtwo}{-\treeoffsetthree}%
\setlength{\branchwidth}{\branchwidthone}%
\ifdim\branchwidthtwo>\branchwidth%
\setlength{\branchwidth}{\branchwidthtwo}\fi%
\setlength{\daughteroffsetone}{\branchwidth}%
\addtolength{\daughteroffsetone}{-\branchwidthone}%
\addtolength{\daughteroffsetone}{-\treeshiftone}%
\setlength{\daughteroffsettwo}{\branchwidth}%
\addtolength{\daughteroffsettwo}{-\branchwidthtwo}%
\addtolength{\daughteroffsettwo}{-\treeshifttwo}%
\setlength{\parentoffset}{-0.5\wd\parentbox}%
\addtolength{\parentoffset}{\treeoffsetthree}%
\addtolength{\parentoffset}{\branchwidth}%
\setlength{\daughteroffset}{0in}%
\ifdim\parentoffset<0in%
\setlength{\daughteroffset}{-\parentoffset}%
\setlength{\parentoffset}{0in}\fi%
\setlength{\parentwidth}{\parentoffset}%
\addtolength{\parentwidth}{\wd\parentbox}%
\setlength{\treeoffset}{\daughteroffset}%
\addtolength{\treeoffset}{\treeoffsetthree}%
\setlength{\treewidth}{\wd\treeboxone}%
\addtolength{\treewidth}{\daughteroffsetone}%
\addtolength{\treewidth}{\treewidthtwo}%
\addtolength{\treewidth}{\daughteroffsettwo}%
\addtolength{\treewidth}{\treewidththree}%
\addtolength{\treewidth}{\daughteroffset}%
\ifdim\treewidth<\parentwidth\setlength{\treewidth}{\parentwidth}\fi%
\sbox{\treebox}{\begin{minipage}{\treewidth}%
\begin{flushleft}%
\hspace*{\parentoffset}\usebox{\parentbox}\\
{\setlength{\unitlength}{0.5\branchwidth}%
\hspace*{\treeoffset}\begin{picture}(4,1)%
\put(0,0){\line(2,1){2}}%
\put(2,0){\line(0,1){1}}%
\put(4,0){\line(-2,1){2}}%
\end{picture}}\\
\vspace{-\baselineskip}
\hspace*{\daughteroffset}%
\makebox[\treewidththree][l]%
{\raisebox{-\ht\treeboxthree}{\usebox{\treeboxthree}}}%
\hspace*{\daughteroffsettwo}%
\makebox[\treewidthtwo][l]%
{\raisebox{-\ht\treeboxtwo}{\usebox{\treeboxtwo}}}%
\hspace*{\daughteroffsetone}%
\raisebox{-\ht\treeboxone}{\usebox{\treeboxone}}%
\end{flushleft}%
\end{minipage}}%
\setlength{\treeoffsetone}{\parentoffset}%
\addtolength{\treeoffsetone}{0.5\wd\parentbox}%
\setlength{\treeshiftone}{0pt}%
\setlength{\treewidthone}{\treewidth}%
\sbox{\treeboxone}{\usebox{\treebox}}\poptree\poptree%
\else\ifnum\value{branchcount}=4\sbox{\parentbox}{\ontop{#2}}%
\setlength{\branchwidthone}{\treewidthtwo}%
\addtolength{\branchwidthone}{\treeoffsetone}%
\addtolength{\branchwidthone}{-\treeshiftone}%
\addtolength{\branchwidthone}{-\treeoffsettwo}%
\setlength{\branchwidthtwo}{\treewidththree}%
\addtolength{\branchwidthtwo}{\treeoffsettwo}%
\addtolength{\branchwidthtwo}{-\treeshifttwo}%
\addtolength{\branchwidthtwo}{-\treeoffsetthree}%
\setlength{\branchwidththree}{\treewidthfour}%
\addtolength{\branchwidththree}{\treeoffsetthree}%
\addtolength{\branchwidththree}{-\treeshiftthree}%
\addtolength{\branchwidththree}{-\treeoffsetfour}%
\setlength{\branchwidth}{\branchwidthone}%
\ifdim\branchwidthtwo>\branchwidth%
\setlength{\branchwidth}{\branchwidthtwo}\fi%
\ifdim\branchwidththree>\branchwidth%
\setlength{\branchwidth}{\branchwidththree}\fi%
\setlength{\daughteroffsetone}{\branchwidth}%
\addtolength{\daughteroffsetone}{-\branchwidthone}%
\addtolength{\daughteroffsetone}{-\treeshiftone}%
\setlength{\daughteroffsettwo}{\branchwidth}%
\addtolength{\daughteroffsettwo}{-\branchwidthtwo}%
\addtolength{\daughteroffsettwo}{-\treeshifttwo}%
\setlength{\daughteroffsetthree}{\branchwidth}%
\addtolength{\daughteroffsetthree}{-\branchwidththree}%
\addtolength{\daughteroffsetthree}{-\treeshiftthree}%
\setlength{\parentoffset}{-0.5\wd\parentbox}%
\addtolength{\parentoffset}{\treeoffsetfour}%
\addtolength{\parentoffset}{1.5\branchwidth}%
\setlength{\daughteroffset}{0in}%
\ifdim\parentoffset<0in%
\setlength{\daughteroffset}{-\parentoffset}%
\setlength{\parentoffset}{0in}\fi%
\setlength{\parentwidth}{\parentoffset}%
\addtolength{\parentwidth}{\wd\parentbox}%
\setlength{\treeoffset}{\daughteroffset}%
\addtolength{\treeoffset}{\treeoffsetfour}%
\setlength{\treewidth}{\wd\treeboxone}%
\addtolength{\treewidth}{\daughteroffsetone}%
\addtolength{\treewidth}{\treewidthtwo}%
\addtolength{\treewidth}{\daughteroffsettwo}%
\addtolength{\treewidth}{\treewidththree}%
\addtolength{\treewidth}{\daughteroffsetthree}%
\addtolength{\treewidth}{\treewidthfour}%
\addtolength{\treewidth}{\daughteroffset}%
\ifdim\treewidth<\parentwidth\setlength{\treewidth}{\parentwidth}\fi%
\sbox{\treebox}{\begin{minipage}{\treewidth}%
\begin{flushleft}%
\hspace*{\parentoffset}\usebox{\parentbox}\\
{\setlength{\unitlength}{0.5\branchwidth}%
\hspace*{\treeoffset}\begin{picture}(6,1)%
\put(0,0){\line(3,1){3}}%
\put(2,0){\line(1,1){1}}%
\put(4,0){\line(-1,1){1}}%
\put(6,0){\line(-3,1){3}}%
\end{picture}}\\
\vspace{-\baselineskip}
\hspace*{\daughteroffset}%
\makebox[\treewidthfour][l]%
{\raisebox{-\ht\treeboxfour}{\usebox{\treeboxfour}}}%
\hspace*{\daughteroffsetthree}%
\makebox[\treewidththree][l]%
{\raisebox{-\ht\treeboxthree}{\usebox{\treeboxthree}}}%
\hspace*{\daughteroffsettwo}%
\makebox[\treewidthtwo][l]%
{\raisebox{-\ht\treeboxtwo}{\usebox{\treeboxtwo}}}%
\hspace*{\daughteroffsetone}%
\raisebox{-\ht\treeboxone}{\usebox{\treeboxone}}%
\end{flushleft}%
\end{minipage}}%
\setlength{\treeoffsetone}{\parentoffset}%
\addtolength{\treeoffsetone}{0.5\wd\parentbox}%
\setlength{\treeshiftone}{0pt}%
\setlength{\treewidthone}{\treewidth}%
\sbox{\treeboxone}{\usebox{\treebox}}\poptree\poptree\poptree%
\else\ifnum\value{branchcount}=5\sbox{\parentbox}{\ontop{#2}}%
\setlength{\branchwidthone}{\treewidthtwo}%
\addtolength{\branchwidthone}{\treeoffsetone}%
\addtolength{\branchwidthone}{-\treeshiftone}%
\addtolength{\branchwidthone}{-\treeoffsettwo}%
\setlength{\branchwidthtwo}{\treewidththree}%
\addtolength{\branchwidthtwo}{\treeoffsettwo}%
\addtolength{\branchwidthtwo}{-\treeshifttwo}%
\addtolength{\branchwidthtwo}{-\treeoffsetthree}%
\setlength{\branchwidththree}{\treewidthfour}%
\addtolength{\branchwidththree}{\treeoffsetthree}%
\addtolength{\branchwidththree}{-\treeshiftthree}%
\addtolength{\branchwidththree}{-\treeoffsetfour}%
\setlength{\branchwidthfour}{\treewidthfive}%
\addtolength{\branchwidthfour}{\treeoffsetfour}%
\addtolength{\branchwidthfour}{-\treeshiftfour}%
\addtolength{\branchwidthfour}{-\treeoffsetfive}%
\setlength{\branchwidth}{\branchwidthone}%
\ifdim\branchwidthtwo>\branchwidth%
\setlength{\branchwidth}{\branchwidthtwo}\fi%
\ifdim\branchwidththree>\branchwidth%
\setlength{\branchwidth}{\branchwidththree}\fi%
\ifdim\branchwidthfour>\branchwidth%
\setlength{\branchwidth}{\branchwidthfour}\fi%
\setlength{\daughteroffsetone}{\branchwidth}%
\addtolength{\daughteroffsetone}{-\branchwidthone}%
\addtolength{\daughteroffsetone}{-\treeshiftone}%
\setlength{\daughteroffsettwo}{\branchwidth}%
\addtolength{\daughteroffsettwo}{-\branchwidthtwo}%
\addtolength{\daughteroffsettwo}{-\treeshifttwo}%
\setlength{\daughteroffsetthree}{\branchwidth}%
\addtolength{\daughteroffsetthree}{-\branchwidththree}%
\addtolength{\daughteroffsetthree}{-\treeshiftthree}%
\setlength{\daughteroffsetfour}{\branchwidth}%
\addtolength{\daughteroffsetfour}{-\branchwidthfour}%
\addtolength{\daughteroffsetfour}{-\treeshiftfour}%
\setlength{\parentoffset}{-0.5\wd\parentbox}%
\addtolength{\parentoffset}{\treeoffsetfive}%
\addtolength{\parentoffset}{2\branchwidth}%
\setlength{\daughteroffset}{0in}%
\ifdim\parentoffset<0in%
\setlength{\daughteroffset}{-\parentoffset}%
\setlength{\parentoffset}{0in}\fi%
\setlength{\parentwidth}{\parentoffset}%
\addtolength{\parentwidth}{\wd\parentbox}%
\setlength{\treeoffset}{\daughteroffset}%
\addtolength{\treeoffset}{\treeoffsetfive}%
\setlength{\treewidth}{\wd\treeboxone}%
\addtolength{\treewidth}{\daughteroffsetone}%
\addtolength{\treewidth}{\treewidthtwo}%
\addtolength{\treewidth}{\daughteroffsettwo}%
\addtolength{\treewidth}{\treewidththree}%
\addtolength{\treewidth}{\daughteroffsetthree}%
\addtolength{\treewidth}{\treewidthfour}%
\addtolength{\treewidth}{\daughteroffsetfour}%
\addtolength{\treewidth}{\treewidthfive}%
\addtolength{\treewidth}{\daughteroffset}%
\ifdim\treewidth<\parentwidth\setlength{\treewidth}{\parentwidth}\fi%
\sbox{\treebox}{\begin{minipage}{\treewidth}%
\begin{flushleft}%
\hspace*{\parentoffset}\usebox{\parentbox}\\
{\setlength{\unitlength}{0.5\branchwidth}%
\hspace*{\treeoffset}\begin{picture}(8,1)%
\put(0,0){\line(4,1){4}}%
\put(2,0){\line(2,1){2}}%
\put(4,0){\line(0,1){1}}%
\put(6,0){\line(-2,1){2}}%
\put(8,0){\line(-4,1){4}}%
\end{picture}}\\
\vspace{-\baselineskip}
\hspace*{\daughteroffset}%
\makebox[\treewidthfive][l]%
{\raisebox{-\ht\treeboxfour}{\usebox{\treeboxfive}}}%
\hspace*{\daughteroffsetfour}%
\makebox[\treewidthfour][l]%
{\raisebox{-\ht\treeboxfour}{\usebox{\treeboxfour}}}%
\hspace*{\daughteroffsetthree}%
\makebox[\treewidththree][l]%
{\raisebox{-\ht\treeboxthree}{\usebox{\treeboxthree}}}%
\hspace*{\daughteroffsettwo}%
\makebox[\treewidthtwo][l]%
{\raisebox{-\ht\treeboxtwo}{\usebox{\treeboxtwo}}}%
\hspace*{\daughteroffsetone}%
\raisebox{-\ht\treeboxone}{\usebox{\treeboxone}}%
\end{flushleft}%
\end{minipage}}%
\setlength{\treeoffsetone}{\parentoffset}%
\addtolength{\treeoffsetone}{0.5\wd\parentbox}%
\setlength{\treeshiftone}{0pt}%
\setlength{\treewidthone}{\treewidth}%
\sbox{\treeboxone}{\usebox{\treebox}}\poptree\poptree\poptree\poptree%
\else\typeout{QobiTeX warning--- Can't handle #1 branching}\fi\fi\fi\fi\fi}
\newcommand{\tree}{%
\usebox{\treeboxone}
\setlength{\treeoffsetone}{\treeoffsettwo}%
\sbox{\treeboxone}{\usebox{\treeboxtwo}}%
\poptree}

\begin{document}

\begin{flushright}
   Published in: {\ems 3e colloque international sur les grammaires 
\\ d'arbres adjoints (TAG+3)} (Technical Report TALANA-RT-94-01), 
\\ eds.\ Anne Abeill{\'e}, Sophie Aslanides and Owen Rambow, 73--76.
\end{flushright}

\begin{center}
   {\bf Adnominal adjectives, code\-switching and lexicalized TAG}

\vspace{\medskipamount}

   Shahrzad Mahootian		    \hfill Beatrice Santorini
\\ Northeastern Illinois University \hfill Northwestern University
\\ usmahoot@uxa.ecn.bgu.edu         \hfill b-santorini@nwu.edu  

September 27, 1994
\end{center}

\nn {\bf 1.  Introduction}

In code\-switching contexts, the language of a syntactic head determines the
distribution of its complements.  Mahootian 1993 
\nocite{mahootian93-diss}
derives this generalization by representing heads as the anchors of
elementary trees in a lexicalized TAG.  However, not all code\-switching
sequences are amenable to a head-complement analysis.  For instance,
adnominal adjectives can occupy positions not available to them in their
own language, and the TAG derivation of such sequences must use unanchored
auxiliary trees.
\bex
\ex\label{abstract-examples}
	\bxl
	\ex[ ]{
	palabras heavy-duty \\
	`heavy-duty words' \\
        (Spanish-English; Poplack 1980:584)}
\nocite{poplack80}
	\ex[ ]{\label{lousy-taste}
        taste lousy sana \\
	`very lousy taste' \\
        (English-Swahili; Myers-Scotton 1993:29, (10))}
\nocite{myers-scotton93-dueling}
	\exl
\eex
Given the null hypothesis that code\-switching and monolingual sequences are
derived in an identical manner, sequences as in (\ref{abstract-examples})
provide evidence that pure lexicalized TAGs are inadequate for the
description of natural language.

\nn {\bf 2.  Heads and complements}

\nn {\bf 2.1  The pivotal role of heads}

Consider code\-switching between a prepositional language like English and
a postpositional one like Hindi.  In such a case, there are four
conceivable switch sequences for an adposition and its object: an English
preposition could precede or follow a Hindi object, and a Hindi
postposition could precede or follow an English object.  Of these four
potential switches, however, only two turn out to be attested---namely
those in which the adposition preserves its language-particular direction
of government.  This is illustrated in (\ref{adposition-good}) and
(\ref{adposition-bad}) (data modified from Pandit 1986:94).
\nocite{pandit86}
\bex
\ex\label{adposition-good}
	\bxl
	\ex[ ]{
	\gll vo hameshaa daftar me {\em on} {\em samay} {aataa hai}. \\
	     he always   office in      {}       time   {comes} \\
	\glt `He always comes to the office on time.'}
	\ex[ ]{he always comes to the office {\ems time par} (`on').}
	\exl
\ex\label{adposition-bad}
	\bxl
	\ex[*]{vo hameshaa daftar me {\ems samay on} aataa hai.}
	\ex[*]{he always comes to the office {\ems par time}.}
	\exl
\eex
Based on a comprehensive cross-linguistic survey of code\-switching data
involving adpositions and other syntactic heads (determiners, verbs,
inflectional morphemes, etc.), Mahootian 1993
\nocite{mahootian93-diss}
proposes the principle in (\ref{key-insight}) to account for the contrast
between (\ref{adposition-good}) and (\ref{adposition-bad}) (see also Pandit
1990:43).
\nocite{pandit90}
\bex
\ex\label{key-insight}
	The language of a syntactic head determines the phrase structure 
	position of its \\
        complements in code\-switching contexts and monolingual contexts alike.
\eex

\nn {\bf 2.2  Heads are anchors}

Mahootian goes on to derive (\ref{key-insight}) by representing syntactic
heads as the anchors of elementary trees in a lexicalized TAG.  The trees
for English {\ems on} and Hindi {\ems par} `on' are given in
(\ref{adposition-trees}a,b).  The attested adpositional phrases in
(\ref{adposition-good}) can be derived simply by substituting an object
from the `other' language at the DP node (`DP' stands for `determiner
phrase' and refers to the traditional noun phrase).  By contrast, deriving
the unattested adpositional phrases in (\ref{adposition-bad}) would require
the trees in (\ref{adposition-trees}c,d).  But since these trees are not
part of the grammars of English and Hindi, there is no derivation for the
code\-switching sequences in (\ref{adposition-bad}), and they are correctly
expected not to occur.
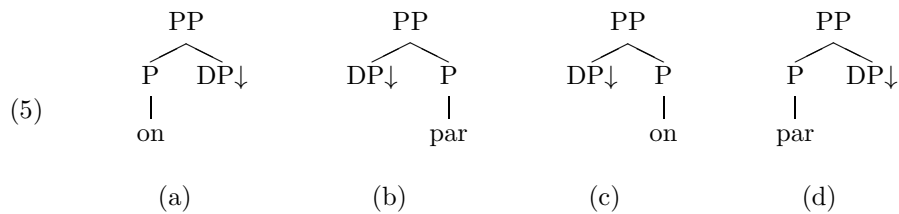
\bex
\ex\label{adposition-trees}
\begin{center}
\begin{tabular}{cccc}
% good `on' tree
	\leaf{on}
	\branch{1}{P}
	\leaf{DP$\downarrow$}
	\branch{2}{PP}
	\tree &
% % good `par' tree
	\leaf{DP$\downarrow$}
 	\leaf{par}
 	\branch{1}{P}
 	\branch{2}{PP}
 	\tree &
% bad `on' tree
	\leaf{DP$\downarrow$}
 	\leaf{on}
 	\branch{1}{P}
	\branch{2}{PP}
 	\tree &
% bad `par' tree
 	\leaf{par}
 	\branch{1}{P}
	\leaf{DP$\downarrow$}
 	\branch{2}{PP}
	\tree \\
& & & \\
(a) & (b) & (c) & (d)
\end{tabular}
\end{center}
\eex

Note that since every clause contains at least one head, this analysis
rules out sequences like (\ref{joshi-sequences}), in which all the
terminals of a clause are from one language, but in an order peculiar to
the other.
\bex
\ex[*]{\label{joshi-sequences}
	He always office to time on comes.}
\eex
The absence of such sequences is noted by Joshi 1985:204, fn.~7,
\nocite{joshi85-cs} 
but is left unresolved there.

\nn {\bf 3.  Adnominal adjectives and nouns}

\nn {\bf 3.1  Adjective-noun sequences are not endocentric structures}

The lexicalized TAG analysis of the distribution of heads and complements
covers much of the relevant code\-switching data.  However, as we will show
in this section, not all code\-switching sequences are instances of switches
between heads and their complements.

It has been suggested a number of times in the code\-switching literature
(Aguirre 1976,
\nocite{aguirre76}
Wentz 1977,
\nocite{wentz77}
Bentahila and Davies 1983)
\nocite{bentahila/davies83}
that the position of an adnominal adjective with regard to the noun it
modifies is determined by the language of the adjective.  Conversely,
Pandit 1990
\nocite{pandit90}
suggests that it is the noun that is the head in such sequences (see also
di~Sciullo, Muysken and Singh 1986).
\nocite{di-sciullo-et-al86} % p.9
Both proposals agree in taking adnominal adjectives and nouns to stand in
the head-complement relationship, but differ as to what is the head and
what is the complement.  Neither of them, however, adequately describes the
full range of code\-switching sequences that is found in connection with
adnominal adjectives.

As in the case of pre- and postpositions, there are four potential
sequences in code\-switching between Adj-N languages like English and N-Adj
languages like Irish, Swahili or the Romance languages: an adjective from
an Adj-N language could precede or follow the noun it modifies, and an
adjective from an N-Adj language could precede or follow the noun it
modifies.  If adnominal adjectives were headed, we would again expect half
of the potential sequences not to occur.  Specifically, if adjective-noun
sequences were headed by adjectives, then (\ref{adj-n-sequences}a,c)
shouldn't occur, whereas if they were headed by nouns, then
(\ref{adj-n-sequences}b,d) shouldn't.  In fact, however, all four sequences
are attested.
\bex
\ex\label{adj-n-sequences}
	\bxl
	\ex[ ]{
	Adjective from N-Adj language, noun from Adj-N language: \\
	I got a lotta {\ems blanquito} (`whitey') {\ems friends}.  \\
	(English-Spanish; Poplack 1980:600, (16b))
	\nocite{poplack80} }
	\ex[ ]{
	Adjective from Adj-N language, noun from N-Adj language: 
	\gll Ma  ci    stanno dei    {\em smart} {\em italiani}.  \\
	     but there are    of-the      {}          Italians \\
	\glt `But there are smart Italians.' \\
		(Italian-English; di Sciullo, Muysken and Singh
		1986:15, (40a)) 
		\nocite{di-sciullo-et-al86} }
\pagebreak
	\ex[ ]{
	Noun from N-Adj language, adjective from Adj-N language:
	\gll T{\'a} {\em carr} {\em light green} aige. \\
	     be          car        {}           at-him \\
	\glt `He has a light green car.' \\
		(Irish-English; Stenson 1990:171, (7a)) 
		\nocite{stenson90} }
	\ex[ ]{
	Noun from Adj-N language, adjective from N-Adj language:  \\
	He presented a {\ems paper exceptionnel} (`exceptional'). \\
	(English-French; Bokamba 1989:282, (16a))
	\nocite{bokamba89} }
	\exl
\eex

It has been suggested to us that the exceptional behavior of adnominal
adjectives might reflect borrowing or imperfect second-language
acquisition.  But since we know of no evidence that borrowing favors
adjectives or that adjectives pose greater difficulties for second-language
acquisition than other syntactic categories, this attempt to explain the
facts in (\ref{adj-n-sequences}) would carry over to the head-complement
case, incorrectly leading us to expect sequences like
(\ref{adposition-bad}).
% Indeed, since code\-switching tokens like those in (\ref{adj-n-sequences})
% are reported quite frequently, despite the low baseline frequency with
% which adnominal adjectives occur in naturally-occurring discourse, the
% proposed explanation would lead us to expect sequences like
% (\ref{adposition-bad}) not just to occur, but to occur frequently, contrary
% to fact.  
As a result, we reject an analysis of the contrast between
(\ref{adposition-good})/(\ref{adposition-bad}) and (\ref{adj-n-sequences})
that relies on extrasyntactic factors.

\nn {\bf 3.2  Adjective-noun sequences as exocentric constructions}

Instead, we interpret the distinct behavior of heads and complements on the
one hand and adnominal adjectives and nouns on the other as evidence that
adnominal adjectives, like other heads, project no further than their
maximal projection (see (\ref{adj-trees}a,b)), but that unlike other
categories, they are introduced into TAG derivations by unanchored
auxiliary trees as in (\ref{adj-trees}c,d).  We will refer to such
unanchored auxiliary trees as modifier trees (= the `athematic' trees of
Kroch 1989).
\nocite{kroch89-asymmetries}
Adj-N languages like English and the related Germanic languages use the
modifier tree in (\ref{adj-trees}c), whereas N-Adj languages like Irish,
Swahili and the Romance languages use the one in (\ref{adj-trees}d).
\bex
\ex
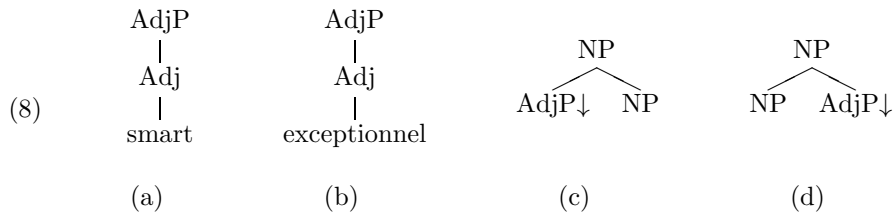
\label{adj-trees}
\begin{center}
\begin{tabular}{cccc}
% smart tree
	\leaf{smart}
	\branch{1}{Adj}
	\branch{1}{AdjP}
	\tree &
% exceptionnel
	\leaf{exceptionnel}
	\branch{1}{Adj}
	\branch{1}{AdjP}
	\tree &
% adj-np tree
	\leaf{AdjP$\downarrow$}
	\leaf{NP}
	\branch{2}{NP}
	\tree &
% np-adj tree
	\leaf{NP}
	\leaf{AdjP$\downarrow$}
	\branch{2}{NP}
	\tree \\
& & & \\
(a) & (b) & (c) & (d)
\end{tabular}
\end{center}
\eex

We would like to emphasize that monolingual data provide no evidence for
introducing adnominal adjectives by using trees like those in
(\ref{adj-trees}) rather than by using anchored trees like those in
(\ref{alt-adj-trees}).
\bex
\ex
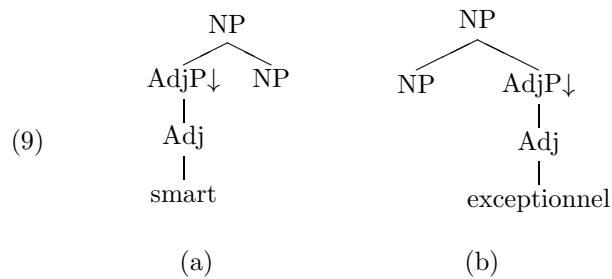
\label{alt-adj-trees}
\begin{center}
\begin{tabular}{cc}
% adj-np tree
	\leaf{smart}
	\branch{1}{Adj}
	\branch{1}{AdjP$\downarrow$}
	\leaf{NP}
	\branch{2}{NP}
	\tree &
% np-adj tree
	\leaf{NP}
	\leaf{exceptionnel}
	\branch{1}{Adj}
	\branch{1}{AdjP$\downarrow$}
	\branch{2}{NP}
	\tree \\
& \\
(a) & (b)
\end{tabular}
\end{center}
\eex
Only in code\-switching contexts does it become apparent that trees like
(\ref{alt-adj-trees}), though possible elementary trees from a formal point
of view, are not elementary structures of the (mental) grammar, but are
instead composed by substituting anchored trees like (\ref{adj-trees}a,b)
at the AdjP nodes of unanchored trees like (\ref{adj-trees}c,d).  It could
of course be argued that adjective-noun sequences in code\-switching
contexts are derived using modifier trees, whereas their monolingual
counterparts are derived using anchored trees.  Such a solution is
reminiscent of the ``third grammar'' approach to code\-switching advocated
by Sankoff and Poplack 1980,
\nocite{sankoff/poplack80}
but roundly rejected as a violation of Occam's razor by Woolford 1983,
\nocite{woolford83}
Joshi 1985,
\nocite{joshi85-cs}
Mahootian 1993,
\nocite{mahootian93-diss}
and Myers-Scotton 1993,
\nocite{myers-scotton93-dueling}
among others.  Following this latter group of authors, we assume the null
hypothesis---namely, that the derivations of code\-switching and
monolingual sequences are formally identical---and we conclude that
monolingual grammars include modifier trees like (\ref{adj-trees}c,d).

% \pagebreak
\nn {\bf 4.  Implications for parsing performance}

Since lexicalization improves the runtime of parsing algorithms, the
question arises whether the necessity of including modifier trees in
grammars for natural language significantly compromises the gains
associated with lexicalization.  Since the set of linguistically motivated
modifier trees is extremely small (adnominal adjectives, manner adverbs,
relative clauses, appositives and perhaps a few more), we expect the
introduction of modifier trees to have no seriously detrimental effect on
parsing performance.  One practical solution, which we will leave for
future research, would be to treat parsing as a two-stage process, in which
the introduction of modifier trees into a derivation must be licensed by an
appropriate anchored tree.

\begin{center}
{\bf Acknowledgments}
\end{center}
\vspace{-0.75\baselineskip}

We are grateful to Stefan Frisch, Lewis Gebhardt, Owen Rambow, B.~Srinivas
and the XTAG group at the University of Pennsylvania for extended
discussions and many helpful comments, and to Rami Nair for last-minute
help with the data in (\ref{adposition-good}) and (\ref{adposition-bad}).
Of course, we alone are responsible for the content of the paper.

\begin{center}
{\bf References}
\end{center}
\vspace{-1.25\baselineskip}

\newcommand{\entry}{\item}
\begin{list}{}{\leftmargin  2.5em
               \itemindent -2.5em
               \itemsep     0pt
               \parsep      0pt
              }

\item 
Aguirre, A. 1976.
Acceptability judgements of code-switching phrases by {C}hicanos:
some preliminary findings.
{ERIC ED} 129 122.

\item
Bentahila, Abdel{\^a}li and Eirlys~E. Davies.
1983.
The syntax of {A}rabic-{F}rench code-switching.
{\em Lingua}\/ 59:301--330.

\item 
Bokamba, Eyamba~G.
1989.
Are there syntactic constraints on code-mixing?
{\em World {E}nglishes}\/ 8:277--293.

\item
{di Sciullo}, Anne-Marie, Pieter Muysken, and Rajendra Singh.
1986.
Government and code-switching.
{\em Journal of linguistics}\/ 22:1--24.

\item
Joshi, Aravind~K. 1985.
Processing of sentences with intra-sentential code-switching.
In David Dowty, Lauri Karttunen, and Arnold Zwicky (eds.), {\em
Natural language processing: psycholinguistic, computational and theoretical
perspectives}\/. New York: Cambridge {U}niversity {P}ress.

\item
Joshi, Aravind~K. and Yves Schabes. 1992.
Tree-{A}djoining {G}rammar and lexicalized grammars.
In Maurice Nivat and Andreas Podelski (eds.), {\em Tree automata and
  languages}\/, 409--431. Elsevier Science.

\item
Kroch, Anthony~S. 1989.
Asymmetries in long distance extraction in a {TAG} grammar.
In Mark Baltin and Anthony~S. Kroch (eds.), {\em Alternative
conceptions of phrase structure}\/, 66--98. Chi\-cago: {U}niversity 
of {C}hicago {P}ress.

\item
Mahootian, Shahrzad. 1993.
{\em A null theory of codeswitching}\/.
PhD thesis, {N}orthwestern {U}niversity.

\item
Myers-Scotton, Carol. 1993.
{\em Duelling languages: grammatical structure in codeswitching}\/.
Oxford: Oxford {U}niversity {P}ress.

\item
Pandit, Ira. 1986.
{\em Hindi {E}nglish code switching. {M}ixed {H}indi {E}nglish}\/.
Delhi: Datta Book Centre.

\item
Pandit, Ira. 1990.
Grammaticality in code switching.
In Rodolfo Jacobson (ed.), {\em Codeswitching as a worldwide
phenomenon}\/, 33--69. New York: Peter Lang.

\item
Poplack, Shana.
1980.
``{S}ometimes {I}'ll start a sentence in {S}panish {\em y termino en
espa\~{n}ol\/}'': toward a typology of code-switching.
{\em Linguistics}\/ 18:581--618.

\item
Sankoff, David and Shana Poplack.
1980.
A formal grammar for code-switching.
{\em Working papers in the {C}enter for {P}uerto {R}ican {S}tudies}\/ 8.

\item
Stenson, Nancy. 1990.
Phrase structure congruence, government, and {I}rish-{E}nglish code
switching.
In Randall Hendrick (ed.), {\em The syntax of the modern {C}eltic
languages}\/, Syntax and semantics 23, 167--197. San Diego: Academic {P}ress.

\item
Wentz, Jim. 1977.
{\em Some considerations in the development of a syntactic
description of code-switching}\/.
PhD thesis, University of Illinois at Urbana-Champaign.

\item
Woolford, Ellen.
1983.
Bilingual code-switching and syntactic theory.
{\em Linguistic inquiry}\/ 14:520--536.

\end{list}

% \pagebreak
% \bibliographystyle{nllt}
% \bibliography{\bib }

\end{document}